\documentclass[11pt,a4paper]{article}
\usepackage[hyperref]{emnlp2020}
\usepackage{times}
\usepackage{latexsym}

\usepackage{amsmath}
\usepackage{amssymb}
\usepackage{bm}
\usepackage{multirow}
\usepackage{graphics}
\usepackage{adjustbox}
\usepackage{arydshln}

\usepackage{microtype}

\aclfinalcopy % Uncomment this line for the final submission
\title{BioMegatron: Larger Biomedical Domain Language Model}

\author{Hoo-Chang Shin, Yang Zhang, Evelina Bakhturina,\\
  \textbf{Raul Puri, Mostofa Patwary, Mohammad Shoeybi, Raghav Mani}
 \\
  NVIDIA / Santa Clara, California, USA \\
  \texttt{hshin@nvidia.com} \\}

\date{}

\begin{document}
\maketitle
\begin{abstract}
There has been an influx of biomedical domain-specific language models, showing language models pre-trained on biomedical text perform better on biomedical domain benchmarks than those trained on general domain text corpora such as Wikipedia and Books.
Yet, most works do not study the factors affecting each domain language application deeply.
Additionally, the study of model size on domain-specific models has been mostly missing.
We empirically study and evaluate several factors that can affect performance on domain language applications, such as the sub-word vocabulary set, model size, pre-training corpus, and domain transfer.
We show consistent improvements on benchmarks with our larger BioMegatron model trained on a larger domain corpus, contributing to our understanding of domain language model applications.
We demonstrate noticeable improvements over the previous state-of-the-art (SOTA) on standard biomedical NLP benchmarks of named entity recognition, relation extraction, and question answering.
Model checkpoints and code are available at \url{ngc.nvidia.com} and \url{github.com/NVIDIA/NeMo}.
% Code and checkpoints to reproduce our experiments are available at \url{github.com/NVIDIA/NeMo}.
% In the last year there has been an influx of biomedical domain-specific language models.
% They show BERT language models pre-trained on biomedical text perform better on bio-domain benchmarks than those trained on Wikipedia and books.
% Yet, the improvements in performance with a BERT-like approach over pre-BERT methods on domain benchmarks appear to be smaller than what has been observed in general NLP benchmarks.
% Most notably, previous findings show the increase in model size often affects performance negatively, contrary to the recent findings in NLP.\\
% We empirically evaluate several factors that can affect performance on domain data, such as pre-training corpus, label bias and subword vocabulary set.
% We show consistent improvements of performance on benchmarks with our larger Bio-Megatron model trained on larger corpus, contributing to our understanding of domain language model applications.
% Noticeable improvements over previous state-of-the-art (SOTA) are demonstrated on a number of standard biomedical NLP benchmarks, including: question answering, named entity recognition, and relation extraction.
% Code and checkpoints to reproduce our experiments are available at [http://masked-for-annonymity.org].
\end{abstract}

\section{Introduction}

Effectively transferring the success of BERT~\citep{devlin2018bert} to the biomedical domain, most notably \citet{lee2019biobert} (BioBERT) and \citet{beltagy2019scibert} (SciBERT) inspired a large number of similar works last year.
For example, \citet{peng2019transfer,alsentzer2019publicly,huang2019clinicalbert} added clinical text to the PubMed biomedical pre-training corpus and tested on standard biomedical and clinical NLP benchmarks.
Many other similar works appeared at the ACL BioNLP Workshop~\citep{demner2019proceedings}.

More recently, \citet{gu2020domain} performed a comprehensive study on the pre-training corpus domain, language model masking method, and adversarial training, benchmarking on a number of different datasets for \textit{token classification}, \textit{sequence classification}, and \textit{sequence regression}.

Compared to the previous works, we perform a more detailed study on \textit{(1)} subword vocabulary, \textit{(2)} labeling method, \textit{(2)} model size, and \textit{(3)} domain transfer, showing gains in \textit{token classification}, \textit{sequence classification}, and \textit{question answering}.

\section{Related Works}

% Motivated by the success of BERT~\citep{devlin2018bert}, 
A prime example of Language Models (LMs) in the biomedical domain is BioBERT~\citep{lee2019biobert}. 
It is a transformer LM pre-trained on the PubMed (\url{www.ncbi.nlm.nih.gov/pubmed}) biomedical text corpus comprised of biomedical literature abstracts.
Their pre-training started from the checkpoint of \citet{devlin2018bert} trained on Wikipedia and Books-Corpus.
Independently, \citet{beltagy2019scibert} (SciBERT) pre-trained BERT from scratch using their vocabulary set on scientific text corpora, including PubMed abstracts and computer science papers.
Both demonstrated increased performance over the previous non-BERT SOTA on biomedical benchmarks, including Named Entity Recognition (NER), Relation Extraction (RE), and Question Answering (QA).
BioBERT and SciBERT report similar results on NER and RE, while only BioBERT report QA results.

They inspired other follow-up works~\citep{alsentzer2019publicly, huang2019clinicalbert,peng2019transfer}, most notably translating their success to the clinical domain, adding the MIMIC-III~\citep{johnson2016mimic} clinical text corpus.
\citet{gu2020domain} (PubMedBERT) used the PubMed full-text for pre-training in addition to the abstracts, and use a domain vocabulary set learned from PubMed corpus.

Meanwhile, they mostly report similar NER and RE tests and results, and only BioBERT reports QA results.
Additionally, most use a $\text{BERT}_\text{Base}$ with 110M parameters.
\citet{peng2019transfer} report slightly improved performance on RE using $\text{BERT}_\text{Large}$ while reporting worse results on NER, compared to $\text{BERT}_\text{Base}$.
These results on biomedical tasks do not benefit from scaling model size to the same degree as standard NLP benchmarks such as GLUE or SQuAD \cite{shoeybi2019megatron,T5}.

% We demonstrate new SOTA results on NER, RE, and QA benchmarks using Megatron-LM~\citep{shoeybi2019megatron} that has parameter size equivalent to $\text{BERT}_\text{Large}$.
% We also test Megatron-LM larger than $\text{BERT}_\text{Large}$ to QA.
% Some empirical analyses are made for each downstream benchmarks on why larger parameter models fail and how to overcome them for improved performance using larger size models.

\section{Language Model Pre-training}

\paragraph{$\text{BERT}_\text{Base \& Large}$}

We compare our models to the pre-trained $\text{BERT}_\text{Base \& Large}$ models of BioBERT~\citep{lee2019biobert} and PubMedBERT~\citep{gu2020domain} ($\text{BERT}_\text{Base}$) for fine-tuning and evaluation.
For QA we use the $\text{BERT}_\text{Large}$ variant of BioBERT following the authors' recommendation.

% We first reproduce the previous studies, pre-training $\text{BERT}_\text{Base}$ on PubMed abstract corpus and evaluating on the downstream benchmark tasks.
% Comparison to $\text{BERT}_\text{Large}$ is made to the previous study by \citet{peng2019transfer}.

\paragraph{BioMegatron}

Megatron-LM~\citep{shoeybi2019megatron} was introduced for efficient model parallel training of large LMs, with up to 8.3B parameters.
\citet{shoeybi2019megatron} showed that rearranging the order of the layer normalization and the residual connections is critical to enabling the scaling of the BERT-style models beyond 336m parameters, and we use the same architecture.

Megatron-LM also used a larger pre-training text corpus, comprised of Wikipedia \citep{devlin2018bert}, CC-Stories \citep{ccstories}, RealNews \citep{grover}, and OpenWebtext \cite{Radford2019GPT2}.
For our LM training, we use the 4.5 billion-word PubMed abstract set and the 1.6 billion-word CC0-licensed Commercial Use Collection of the PMC full-text corpus (\url{www.ncbi.nlm.nih.gov/pmc}).

% We use and compare Megatron-LM which has a number of parameters equal to or larger than the $\text{BERT}_\text{Large}$.
% Most previous works pre-train BERT on 4.5 billion words PubMed abstracts, where ~\citet{peng2019transfer} use additional 2.8 billion words MIMIC-III dataset.
% We add 1.6 billion words Commercial Use Collection of PMC\footnote{\url{www.ncbi.nlm.nih.gov/pmc}} full-text corpus, that has ``CC0'' (Creative Commons public domain) license, to the pre-training.

% We test our Bio-Megatron on downstream NER, RE, and QA benchmarks.

We train three sizes of BioMegatron: with 345 million, 800 million, and 1.2 billion number of parameters (Table~\ref{tab:model_size}).
We compare four pre-training scenarios in the smallest 345m model - using BERT-cased/uncased vocabularies, each pre-trained from scratch and fine-tuned from general domain LM.
We also compare two sets of domain vocabularies learned on PubMed text corpus using \texttt{SentencePiece} (github.com/google/sentencepiece) library, each containing 30k and 50k subword units.

We train the larger BioMegatron models with less variation: 800m models from scratch on PubMed with BERT -cased/-uncased vocabularies; and 1.2b model starting from general domain LM checkpoint using BERT-uncased vocabulary.

\begin{table}
\footnotesize
\begin{center}
\resizebox{1\columnwidth}{!}{
\begin{tabular}{c|c|c|c} \hline
\#Parameters & \#Layers & \#Hidden Size & \#Attention Heads \\ \hline
345m      & 24     & 1024   & 16          \\ 
800m      & 36     & 1280   & 20          \\ 
1.2b      & 24     & 2048   & 16         \\ \hline 
\end{tabular}}
\caption{\label{tab:model_size}Model configurations.}
\end{center}
\end{table}

\section{Downstream Benchmark Tasks}

We use the most widely used downstream biomedical benchmark datasets for NER, RE, and QA.

\paragraph{Named Entity Recognition}

The BC5CDR~\citep{li2016biocreative} NER dataset annotated \textit{disease} and \textit{chemical} terms with IOB tagging~\citep{ramshaw1999text}.
In NCBI-disease~\citep{dougan2014ncbi}, only \textit{disease} entities are IOB-tagged.

\paragraph{Relation Extraction}

The ChemProt~\citep{krallinger2015chemdner} %is the most widely used for biomedical relation extraction benchmark. The
dataset contains sentences from PubMed abstracts, where chemical-protein interaction types are annotated as five categories. % (CPR: [3, 4, 5, 6, 9]).
% Drug interactions extracted from PubMed abtracts are annotated into five categories in the DDI~\citep{herrero2013ddi} dataset.
Relation Extraction is essentially a sequence classification task, classifying a set of sentences into a category.
% Essentially the task can be regarded as a sentence/text classification task, classifying a set of sentences into a category.

\paragraph{Question Answering}

The BioASQ-7b factoid task~\citep{tsatsaronis2015overview} is a biomedical QA dataset whose format is similar to the SQuAD dataset~\citep{rajpurkar2016squad}.
In this task, context-snippet, question and answer triplets, and factoid question/answers are evaluated with \textit{strict accuracy (SAcc)}, \textit{lenient accuracy (LAcc)}, and \textit{mean reciprocal rank (MRR)}.

\section{Results and Discussion}

The evaluation results on NER and RE are shown in Table~\ref{final-evaluation-ner-re}, and QA are shown in Table~\ref{final-evaluation-qa}.
We perform entity-level F1 NER using the official CoNLL evaluation script translated into Python (\url{github.com/spyysalo/conlleval.py}). RE uses micro-level F1, and QA uses the BioASQ evaluation script (\url{github.com/BioASQ/Evaluation-Measures}).

%%%%%%%%%%%%%%%%%%%%%%%%%%%%%%%%
\begin{table*}[t]
\centering
\resizebox{0.9\textwidth}{!}{
\begin{tabular}{lllclcccc}
\hline
 & \textbf{Benchmark} & \textbf{Model} & \textbf{\#Parameters} & \textbf{Vocabulary} & \textbf{Prec}&\textbf{Rec}&\textbf{F1} \\ \hline
\multirow{18}{*}{\rotatebox{90}{\textbf{NER}}} 
 & \multirow{6}{*}{BC5CDR-chem} & BioBERT & 110m & BERT-cased & 90.0 & 93.4 & 91.7 \\
                             & & PubMedBERT & 110m & PubMedBERT-vocab (30k) & 92.1 & 93.2 & 92.6 \\
                             & & \textbf{BioMegatron} & \textbf{345m} & \textbf{Bio-vocab-30k} & 92.1 & \textbf{93.6} & \textbf{92.9} \\
                             & & BioMegatron & 345m & Bio-vocab-50k & \textbf{92.9} & 92.0 & 92.5 \\
                             & & BioMegatron & 800m & BERT-cased & 91.3 & 92.9 & 92.1 \\
                             & & BioMegatron & 1.2b & BERT-uncased & 92.0 & 90.5 & 91.3 \\
                             \cdashline{2-8} 
                              
 & \multirow{6}{*}{BC5CDR-disease} & BioBERT & 110m & BERT-cased & 85.0 & 89.4 & 87.2 \\
                             & & PubMedBERT & 110m & PubMedBERT-uncased (30k) & \textbf{86.2} & 88.4 & 87.3 \\
                             & & BioMegatron & 345m & Bio-vocab-30k & 85.2 & 88.8 & 87.0 \\
                             & & \textbf{BioMegatron} & \textbf{345m} & \textbf{Bio-vocab-50k} & 86.1 & \textbf{91.0} & \textbf{88.5} \\
                             & & BioMegatron & 800m & BERT-cased & 85.8 & 90.1 & 87.9 \\
                             & & BioMegatron & 1.2b & BERT-uncased & 83.8 & 89.2 & 86.4 \\
                             \cdashline{2-8} 
                             
 & \multirow{6}{*}{NCBI-disease} & BioBERT & 110m & BERT-cased & 85.0 & 90.0 & 87.5 \\
                             & & PubMedBERT & 110m & PubMedBERT-uncased (30k) & 85.9 & 87.7 & 86.8 \\
                             & & BioMegatron & 345m & Bio-vocab-30k & 85.6 & 88.6 & 87.1 \\
                             & & BioMegatron & 345m & Bio-vocab-50k & 83.7 & \textbf{90.4} & 87.0 \\
                             & & \textbf{BioMegatron} & \textbf{800m} & \textbf{BERT-cased} & \textbf{87.0} & 88.8 & \textbf{87.8} \\
                             & & BioMegatron & 1.2b & BERT-uncased & 83.5 & 90.1 & 86.7 \\
                             
                             \hline

\multirow{6}{*}{\rotatebox{90}{\textbf{RE}}}
 & \multirow{6}{*}{ChemProt} & BioBERT & 110m & BERT-cased & 76.5 & 73.3 & 74.8 \\
                             & & PubMedBERT & 110m & PubMedBERT-uncased (30k) & 73.6 & 77.7 & 75.6 \\
                             & & BioMegatron & 345m & Bio-vocab-30k & 77.8 & 72.5 & 75.1 \\
                             & & \textbf{BioMegatron} & \textbf{345m} & \textbf{Bio-vocab-50k} & 74.5 & \textbf{79.7} & \textbf{77.0} \\
                             & & BioMegatron & 800m & BERT-cased & 80.4 & 68.9 & 74.3 \\
                             & & BioMegatron & 1.2b & BERT-uncased & \textbf{82.0} & 65.6 & 72.9 \\
                             \hline
\end{tabular}}
\caption{\label{final-evaluation-ner-re}Evaluation results on NER and RE after fine-tuning for 30 epochs with hyper-parameter settings of: \texttt{num-fc-layers}: \{1, 2\}; \texttt{fc-hidden-size}: \{512, 1024\}; \texttt{fc-dropout}: 0.5; \texttt{max-seq-length}: 128; \texttt{learning-rate}: 5e-5; cross-entropy loss, with Adam optimizer. BioMegatron models are pre-trained from scratch on PubMed, except 1.2b model which is fine-tuned from a general domain model checkpoint.}
\end{table*}

\begin{table*}[ht!]
\centering
\resizebox{0.8\textwidth}{!}{
\begin{tabular}{lllclcccc}
\hline
 & \textbf{Benchmark} & \textbf{Model} & \textbf{\#Parameters} & \textbf{Vocabulary} & \textbf{SAcc}&\textbf{LAcc}&\textbf{MRR} \\ \hline
\multirow{5}{*}{\rotatebox{90}{\textbf{QA}}} 
 & \multirow{5}{*}{BioASQ-7b-factoid} & BioBERT-Base & 110m & BERT-cased & 30.8 & 64.1 & 41.1 \\
                             & & BioBERT-Large & 345m & BERT-cased & 42.8 & 62.8 & 50.1 \\
                             & & \textbf{BioMegatron} & \textbf{345m} & \textbf{BERT-uncased} & 46.2 & \textbf{62.6} & \textbf{52.5} \\
                             & & BioMegatron & 800m & BERT-uncased & 45.2 & 58.6 & 50.4 \\
                             & & BioMegatron & 1.2b & BERT-uncased & \textbf{47.4} & 60.9 & 52.4 \\
                             \hline
\end{tabular}}
\caption{\label{final-evaluation-qa}Evaluation results on QA after fine-tuning for 30 epochs on checkpoints fine-tuned on SQuAD dataset with fixed hyper-parameter settings as \texttt{num-fc-layers}: 2; \texttt{fc-hidden-size}: 2048; \texttt{fc-dropout}: 0.1; \texttt{max-seq-length}: 512; \texttt{learning-rate}: 3e-5; cross-entropy loss, using Adam optimizer. BioMegatron models are pre-trained from scratch on PubMed, except 1.2b model which is fine-tuned from a general domain model checkpoint.}
\end{table*}

%%%%%%%%%%%%%%%%%%%%%%%%%%%%%

\subsection{Named Entity Recognition}

\begin{figure}[h]
\centering
\includegraphics[width=1\linewidth]{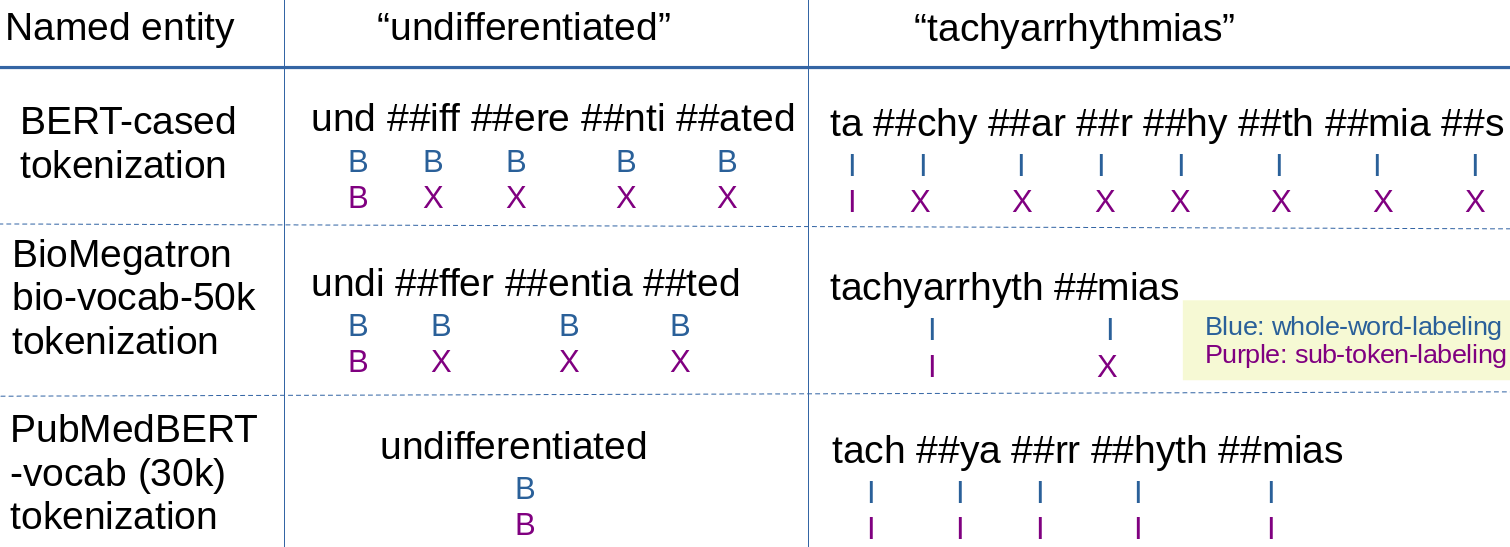}
\caption{Examples of tokenization with different sub-word vocabularies. Blue and purple text show word-level and subtoken-level entity labeling, respectively.}
\label{fig:ner_digin}
\end{figure}

While the NER benchmark datasets appear saturated due to the small sample size, we find that the subword vocabulary is the most critical factor.
Examples of tokenization with different vocabularies are shown in Figure~\ref{fig:ner_digin}.
Representing named entities as single terms is more helpful than breaking them into several subtokens.
Table~\ref{word-breakout-rate} shows the rate named entities break into sub-tokens for each benchmark training set with different sub-word vocabularies.
PubMedBERT vocabulary set has a low break-out rate while being smaller in size than our 50k-size vocabulary.
A lower break-out rate with smaller vocabulary size probably helps achieve better NER performance despite smaller model size.

\begin{table}
\centering
\resizebox{1\columnwidth}{!}{
\begin{tabular}{lccc}
\hline 
\textbf{Sub-word vocabulary} & \textbf{BC5-chem} & \textbf{BC5-disease} \\ \hline
BERT-cased     &       3.012          &         2.42 \\
PubMedBERT-uncased \underline{(30k)} &  \underline{1.654} &  \underline{1.236} \\
BioMegatron-bio-30k-cased &  1.753          &         1.272 \\
\textbf{BioMegatron-bio-50k-cased} &  \textbf{1.478}          &         \textbf{1.116} \\
\hline
\end{tabular}}
\caption{\label{word-breakout-rate}The rate of named entities breaking into subtokens (\#tokens/\#words) in NER training sets.}
\end{table}

We can label the entities for NER training as: \textit{(1)} marking the whole entity as a single label, and \textit{(2)} labeling sub-tokens separately.
Figure~\ref{fig:ner_digin} shows examples of the labeling methods.
We find these different schemes can result in as much as $\sim$2\% difference in the F1-score on NER evaluation, possibly indicating that the datasets are too small.
We report NER results by labeling sub-tokens separately, except for NCBI-disease dataset, which results in better whole-entity labeling across models.

% Larger models tend to provide higher precision in exchange for lower recall.
% Better results with larger models could be obtained with further hyper-parameter tuning and longer training.

\subsection{Relation Extraction}

Since RE is a classification task, albeit on sequences rather than on tokens, the choice of sub-word vocabulary has a notable effect.

We can also observe that larger models result in higher precision for lower recall, both for NER and RE.
More hyper-parameter tuning could achieve higher F1-scores, even the generalization ability of such result may be questionable.

\subsection{Question Answering}

Table~\ref{final-evaluation-qa} show evaluation results after fine-tuning on SQuAD for 10 epochs and BioASQ for 30 epochs each, following the recipe found to work best by \citet{lee2019biobert}.
We found large batch size to be beneficial, as Q\&A pairs repeat up to 88 times.
We use batch size of 64 per GPU with data parallelism on 16 GPUs.
Using biomedical vocabularies result in much worse results, possibly due to its low relevance in the first SQuAD fine-tuning task.

Larger models tend to perform better in QA, though it levels off after 345m parameters.
The larger model size effect is more evident when fine-tuning on BioASQ directly, as shown in Table~\ref{bioasq-finetuning-scratch}.

\begin{table}[h]
\centering
\resizebox{.8\columnwidth}{!}{
\begin{tabular}{lccc}
\hline 
\textbf{Model} & \textbf{SAcc} & \textbf{LAcc} & \textbf{MRR} \\ \hline
BioMegatron-345m &   33.1          &         50.4 & 39.8 \\
BioMegatron-800m &   37.7          &         \textbf{56.3} & 45.1 \\
\textbf{BioMegatron-1.2b} &  \textbf{40.6} & 53.7 & \textbf{45.6} \\
\hline
\end{tabular}}
\caption{\label{bioasq-finetuning-scratch}Results on BioASQ-7b factoid, without fine-tuning on SQuAD dataset first. The other models, including those using domain vocabularies, could not achieve any comparable results. A consistent pattern of improvement over model size noticeable on par with findings in general domain LM on SQuAD.}
\end{table}

\subsection{Domain Transfer and Generalization}

We examine how well a general- or domain- specific LM generalizes across domains related to the model size.
\citet{gu2020domain} studied the effect of ``domain-specific'' vs. ``mixed-domain'' pre-training, i.e., pre-training on PubMed from scratch vs. pre-training starting from a general domain LM (fine-tuning).
They found that pre-training on PubMed from scratch is better for biomedical NLP benchmarks, but we analyze its effect with further pre-training (fine-tuning) steps.
In other words, if starting from a general domain LM, does sufficient fine-tuning make it as good as a fully domain-specific model?
Can such model have any advantage for cross-domain or cross-discipline generalization?

\begin{table}[h]
\centering
\resizebox{0.9\columnwidth}{!}{
\begin{tabular}{lllc}
\hline
 & \textbf{Benchmark} & \textbf{Fine-tuning steps} & \textbf{F1} \\ \hline
\multirow{14}{*}{\rotatebox{90}{NER}} 
 & \multirow{7}{*}{BC5CDR-chem} & $10^3$ steps & 63.2 \\
                                & & $10^4$ steps & 74.3 \\
                                & & $10^5$ steps & 89.7 \\ 
                                & & $2\cdot10^5$ steps & 89.37 \\
                                & & $3\cdot10^5$ steps & 91.8 \\
                                & & $\bm{4\cdot10^5}$ \textbf{steps} & \textbf{92.1} \\
                                & & $5\cdot10^5$ steps & 91.2 \\
                                \cdashline{2-4}
                              
 & \multirow{7}{*}{BC5CDR-disease} & $10^3$ steps & 39.4 \\
                                & & $10^4$ steps & 63.6 \\
                                & & $10^5$ steps & 79.8 \\ 
                                & & $2\cdot10^5$ steps & 81.2 \\
                                & & $3\cdot10^5$ steps & 79.2 \\
                                & & $\bm{4\cdot10^5}$ \textbf{steps} & \textbf{81.9} \\
                                & & $5\cdot10^5$ steps & 81.8 \\
                             \hline

\multirow{7}{*}{\rotatebox{90}{RE}} & \multirow{7}{*}{ChemProt} & $10^3$ steps & 0.00 \\
                                & & $10^4$ steps & 34.1 \\
                                & & $10^5$ steps & 63.4 \\ 
                                & & $\bm{2\cdot10^5}$ \textbf{steps} & \textbf{71.1} \\
                                & & $3\cdot10^5$ steps & 70.4 \\
                                & & $4\cdot10^5$ steps & 69.7 \\
                                & & $5\cdot10^5$ steps & 68.3 \\
                             \hline
\end{tabular}}
\caption{\label{megatron1.2b-ft-ner-re}Comparison of fine-tuning steps for NER and RE benchmark when pre-training general-domain Megatron-1.2b model on PubMed. Cross-domain LMs should be trained sufficiently long on domain text to achieve comparable performance.}
\end{table}

Table~\ref{megatron1.2b-ft-ner-re} shows F1-score evaluation on NER and RE benchmarks using a general-domain BioMegatron-1.2b with additional fine tuning.
It shows that even for a large LM that was pre-trained on a large text corpus, it needs sufficient further pre-training on domain text (PubMed).
After sufficient pre-training on domain text, it can be as good as an LM pre-trained on domain-text only, except that vocabulary has more significant effect on NER.

\begin{table}[h]
\centering
\resizebox{1\columnwidth}{!}{
\begin{tabular}{lccc}
\hline 
\textbf{Model} & \textbf{SAcc} & \textbf{LAcc} & \textbf{MRR} \\ \hline
\textbf{Megatron-345m} (general LM) &   \textbf{38.5} & \textbf{52.6} & \textbf{43.7} \\
Megatron-1.2b (general LM) &   29.3          &         39.7 & 32.7 \\
\hline
\end{tabular}}
\caption{\label{bioasq-general-domain}Fine-tuning and evaluating on BioASQ-7b using general domain LMs not trained on PubMed corpus. Larger model does not perform better.}
\end{table}

\begin{table}[h]
\centering
\resizebox{1\columnwidth}{!}{
\begin{tabular}{lcc}
\hline 
\textbf{Model} & \textbf{SQuAD-v1.1} & \textbf{SQuAD-v2.0}  \\ \hline
BioMegatron-345m &  90.4 & 84.2 \\
BioMegatron-345m-ft & 86.5 & 77.9  \\
BioMegatron-800m & 91.6 & 86.1 \\
\underline{BioMegatron-1.2b-ft} & \underline{91.8} & \underline{86.4} \\
$\text{BERT}_\text{LARGE}$ & 90.9 & 81.8 \\
RoBERTa & 94.6 & 89.4 \\
\textbf{Megatron-3.9b} & \textbf{95.8} & \textbf{91.2} \\
\hline
\end{tabular}}
\caption{\label{squad-evals}Fine-tuning on SQuAD -v1.1/-v2.0 using BioMegatron and evaluating on F1-score on dev-set. BioMegatron with `-ft' are pre-trained from general domain checkpoints (fine-tuned). Results of other general domain LMs are compared: RoBERTa~\citep{liu2019roberta}, Megatron-LM~\citep{shoeybi2019megatron}.}
\end{table}

Table~\ref{bioasq-general-domain} shows the results of general-domain LMs fine-tuned on BioASQ-7b-factoid.
Larger models do not perform better, which may indicate overfitting is occuring on the small training set.

Table~\ref{squad-evals} shows the generalization ability of BioMegatron models on SQuAD datasets.
Here, a large biomedical LM pre-trained on large text corpus performs better than smaller general domain LMs such as $\text{BERT}_\text{LARGE}$, even when pre-trained on the biomedical text.

\subsection{Other Domain-Specific Factors}

\paragraph{Size and Bias in Biomedical Datasets}

Annotating biomedical data requires in-depth domain knowledge.
Besides, data often have substantial label bias as the occurrences of ``abnormal'' or ``findings'' are rare by nature.
As a result, biomedical benchmark data tend to be smaller and highly biased than their general domain counterparts.

\begin{table}[h]
\centering
\resizebox{0.8\columnwidth}{!}{
\begin{tabular}{llrc}
\hline \textbf{Task} & \textbf{Dataset} & \textbf{\# Samples} & \textbf{Bias \%} \\ \hline
\multirow{2}{*}{\textbf{NER}} & CONLL-2003 & 14987 & 0.18 \\
%& NCBI-disease & 5425 & 0.09 \\
& BC5CDR & 5235 & 0.08 \\
\hline
\multirow{2}{*}{\textbf{CLS}} & MRPC & 3668 & 0.48 \\
& ChemProt & 19461 & 0.27 \\ \hline
\multirow{2}{*}{\textbf{QA}} & SQuAD-v1.0 & 87599 & 0.4 \\
& BioASQ-7b & 5537 & 0.02 \\ \hline
\end{tabular}}
\caption{\label{bio-dataset}Label bias in general and biomedical benchmark dataset. CONLL-2003~\citep{sang2003introduction}, MRPC~\citep{dolan2005microsoft}, and SQuAD~\citep{rajpurkar2016squad} are general domain dataset for NER, CLS (RE), and QA, respectively, for comparison against biomedical domain dataset. Label bias is computed as [\textit{sum of the \#samples of minority labels}]/[\textit{\#samples of majority label}], for NER and RE (CLS), and [\textit{\#minimum repeat of the same answer}]/[\textit{\#maximum repeat of the same answer}] for QA.}
\end{table}

Table~\ref{bio-dataset} shows a comparison of benchmark datasets for NER, RE (CLS), and QA in the biomedical domain and their general-domain counterparts.
% Bias is computed as [\textit{sum of the \#samples of minority labels}] divided by [\textit{\#samples of majority label}], for NER and RE (CLS).
% For QA it is computed as [\textit{\#minimum repeat of the same answer}] divided by [\textit{\#maximum repeat of the same answer}].
The SQuAD Q\&A set is 15 times larger than the BioASQ data, where the same question-answer combinations appear up to 88 times in BioASQ.
Question-answer pairs are seldom repeated in SQuAD data, at most twice.
The BC5CDR NER dataset is 1/3 size of CONLL-2003 and the ratio of \texttt{I/O} to \texttt{O} tags 0.08, compared to 0.18 for CONLL.

Methods to circumvent data imbalance issues such as oversampling the minority classes~\citep{chawla2002smote,chen2010ramoboost} and using weighted cross-entropy gave minor effects on our NER and RE benchmarks.
Recently, \citet{li2019dice} proposed dice-loss for data-imbalance issues in NLP, with SOTA results on NER and QA, which could be a future avenue to explore for domain LMs.
Transfer learning showed effectiveness in the biomedical QA task.
However, it is somewhat unclear how to apply it to NER and RE tasks.

\paragraph{Pre-training Corpus and Duration}

\begin{table}[h]
\centering
\resizebox{1\columnwidth}{!}{
\begin{tabular}{llc}
\hline 
\textbf{Model} & \textbf{PubMed Corpus} & \textbf{\#Words}  \\ \hline
BioBERT & abstracts & 4.5 billion \\
PubMedBERT & abstracts + full-text & 16.8 billion \\
BioMegatron & abstracts + full-text-CC & 6.1 billion \\ 
\hline
\end{tabular}}
\caption{\label{pre-training-corpus}Pre-training text corpus of each biomedical LM. We pre-train on PubMed abstracts and full-text commercial-collection (CC) that are free of copyrights.}
\end{table}

PubMedBERT is pre-trained on a much larger text corpus, as shown in Table~\ref{pre-training-corpus}.
It is a performant domain-LM with a larger pre-training corpus and adequate domain vocabulary compared to its model size.
We pre-train our LMs for about one epoch, reaching a masked-LM loss of about 1.2~\citep{devlin2018bert}.
Further pre-training may be helpful, but it is challenging to have strictly controlled experiments with many different settings.

\section{Conclusion}

We review and test several factors that can affect the performance of domain language models.
We find that a language model targeted for a domain and application performs best.
For example, model size is a secondary factor to vocabulary set for token classification task.
Larger model size does not necessarily translate to better performance on a cross-domain benchmark task.

This probably indicates that there is no master model that can ``do it all'', at least well enough as a targeted one.
The model size is a secondary factor; larger model size can probably further improve the performance of a a domain- and application- specific language model.

\section*{Acknowledgement}
The authors would like to thank Sun Kim at NIH/NCBI (now at Amazon Alexa AI) for helpful discussions and suggestions.

\bibliographystyle{acl_natbib}
\bibliography{anthology,emnlp2020}

\end{document}